\documentclass[sigconf]{acmart}
\usepackage{booktabs} 
\usepackage{subfigure}

\usepackage{epsfig}
\usepackage{graphicx}
\usepackage{amsmath}
\usepackage{amssymb}
\usepackage{multirow}

\setcopyright{rightsretained}

\DeclareMathOperator{\conv}{Conv}

\acmConference[MM'18]{ACM Multi Media conference}{October 2018}{Seoul, Korea}
\acmSubmissionID{467}

\begin{document}
\title{Non-locally Enhanced Encoder-Decoder Network for Single Image De-raining 
}


\author{Guanbin Li}
\orcid{0000-0002-4499-3863}
\affiliation{%
  \institution{Sun Yat-sen University}
}
\email{liguanbin@mail.sysu.edu.cn}

\author{Xiang He}
\affiliation{%
  \institution{Sun Yat-sen University}
}
\email{hexiang7@mail2.sysu.edu.cn}

\author{Wei Zhang}
\affiliation{%
  \institution{Sun Yat-sen University}
}
\email{zhangwei.hi@gmail.com}

\author{Huiyou Chang}
\affiliation{%
  \institution{Sun Yat-sen University}
}
\email{isschy@mail.sysu.edu.cn}

\author{Le Dong}
\authornote{Corresponding author is Le Dong. This work was supported by the National Natural Science Foundation of China under Grant 61702565 and Grant 61772114, the Science and Technology Planning Project of Guangdong Province under Grant 2017B010116001, Guangdong Natural Science Foundation Project for Research Teams under Grant 2017A030312006, and was also sponsored by CCF-Tencent Open Research Fund.}

\affiliation{%
  \institution{University of Electronic Science and Technology of China}
}
\email{ledong@uestc.edu.cn}

\author{Liang Lin}
\affiliation{%
  \institution{Sun Yat-sen University}
}
\email{linliang@ieee.org}

\begin{abstract}
Single image rain streaks removal has recently witnessed substantial progress due to the development of deep convolutional neural networks. However, existing deep learning based methods either focus on the entrance and exit of the network by decomposing the input image into high and low frequency information and employing residual learning to reduce the mapping range, or focus on the introduction of cascaded learning scheme to decompose the task of rain streaks removal into multi-stages. These methods treat the convolutional neural network as an encapsulated end-to-end mapping module without deepening into the rationality and superiority of neural network design. In this paper, we delve into an effective end-to-end neural network structure for stronger feature expression and spatial correlation learning. Specifically, we propose a non-locally enhanced encoder-decoder network framework, which consists of a pooling indices embedded encoder-decoder network to efficiently learn increasingly abstract feature representation for more accurate rain streaks modeling while perfectly preserving the image detail. The proposed encoder-decoder framework is composed of a series of non-locally enhanced dense blocks that are designed to not only fully exploit hierarchical features from all the convolutional layers but also well capture the long-distance dependencies and structural information. Extensive experiments on synthetic and real datasets demonstrate that the proposed method can effectively remove rain-streaks on rainy image of various densities while well preserving the image details, which achieves significant improvements over the recent state-of-the-art methods.

\end{abstract}
%
%


\copyrightyear{2018} 
\acmYear{2018} 
\setcopyright{acmcopyright}
\acmConference[MM '18]{2018 ACM Multimedia Conference}{October 22--26, 2018}{Seoul, Republic of Korea}
\acmBooktitle{2018 ACM Multimedia Conference (MM '18), October 22--26, 2018, Seoul, Republic of Korea}
\acmPrice{15.00}
\acmDOI{10.1145/3240508.3240636}
\acmISBN{978-1-4503-5665-7/18/10}

\maketitle

\section{Introduction}
Images with rain streaks are often captured  by outdoor surveillance equipments, which may significantly degrade the performance of some existing computer vision systems and may also result in a pool visual experience for some multimedia applications. Automatic rain streaks removal has thus become a crucial research task in the field of computer vision and multimedia, and has been successfully applied in the fields of driverless technology~\cite{kurihata2005rainy,vasile2010rain} and content based image editing~\cite{huang2012context,shen2011fast,xue2012motion,zhang2006rain}.

\begin{figure}[t]
\begin{center}
   \includegraphics[width=0.48\textwidth]{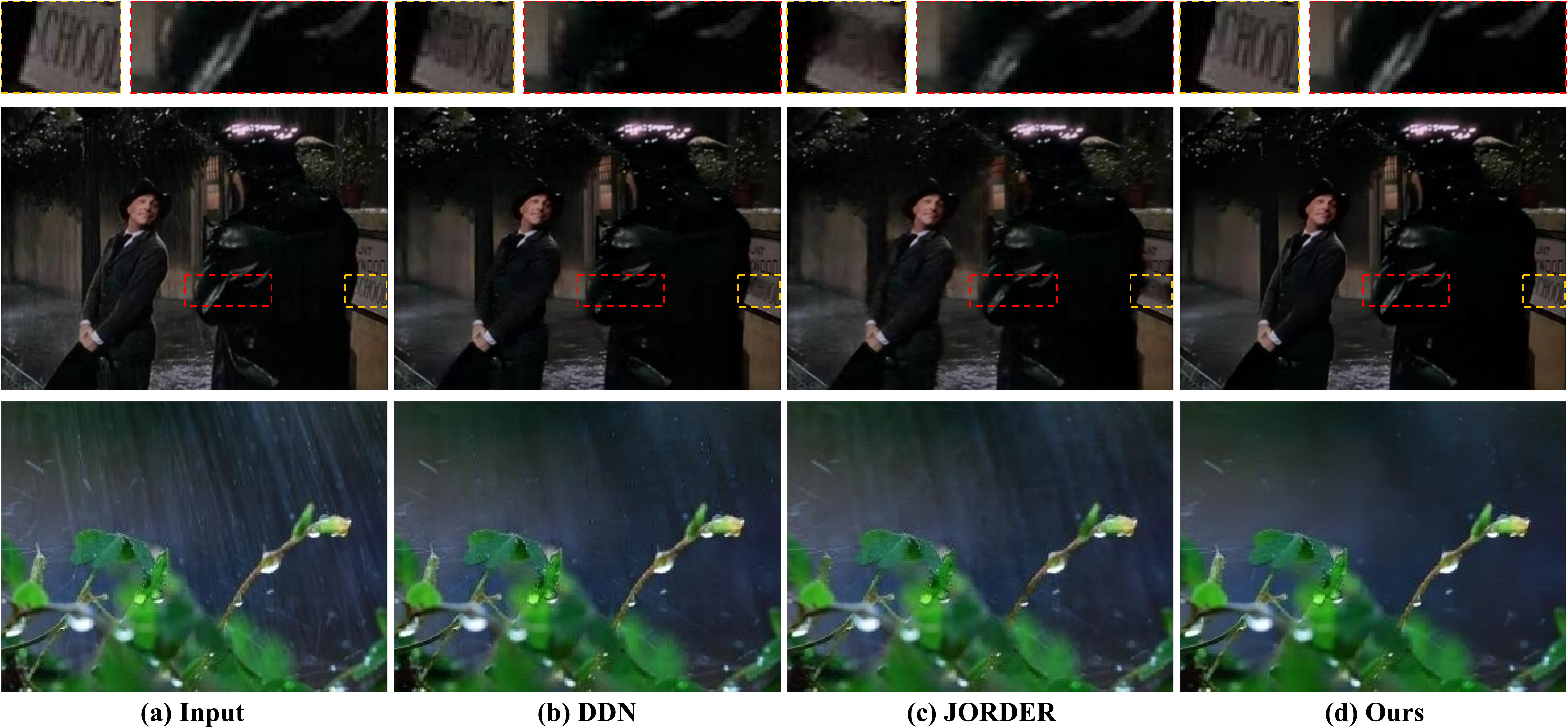}
\end{center}

   \caption{Sample examples of single image de-raining results. (a) Input images with rain-streaks. (b) Results of DDN~\cite{ddn} (c) Results of JORDER~\cite{JRDR} (d) Our results. The first row is the enlargement of the selected regions of the second row, which shows the advantage of our proposed NLEDN in detail preserving. The third row demonstrates the promising result of our NLEDN in removing long rain streaks.}
\label{fig:teaser}
\end{figure}

The research on visual de-raining can be traced back to the last decade. Most of the early research focused on the removal of rain streaks in video sequences captured with static cameras~\cite{video_derain_2,santhaseelan2015utilizing,video_derain_3,garg2005does,video_derain_1,zhang2006rain}. They mostly attempted to solve the problem by exploiting the temporal correlation in the luminance domain  between successive frames~\cite{video_derain_2,video_derain_1,zhang2006rain}. Due to the lack of temporal information, de-raining on single image is more ill-posed, however, it has received widespread research attention due to its greater practicality and challenge~\cite{sparse_coding_1,DID-MDN,JRDR,de-rain_nonlocal_mean,huang2012context}. Traditional methods on single image de-raining explore certain prior information on physical characteristics of rain streaks and model it as a signal separation problem~\cite{low_rank_1,sparse_coding_1,sun2014exploiting,sparse_coding_3}, or directly regard it as an image filtering problem and solve by resorting to nonlocal mean smoothing~\cite{de-rain_nonlocal_mean}. However, since these models are based on handcrafted low-level feature and fixed a priori rain streaks assumptions, they can only cope with rain drops of specific shapes, scales and density, and can easily lead to the destruction of image details which are similar to rain streaks. 

In recent years, due to the powerful feature representation and end-to-end data inference capabilities, deep convolutional neural networks have been widely applied to single image de-raining and have achieved significant performance improvement. These methods generally model the problem as a pixel-wise image regression process which directly learns to map an input rainy image to its clean version or a negative residual map in an end-to-end mode through a series of convolution, pooling, and non-linear operations, etc.  Although considerable progress has been made in comparison with traditional methods, existing deep models still suffer from several limitations. Firstly, most of the deep CNN based models emulate the experience of low-level image processing such as image denoising, super-resolution and filtering, design shallow neural network structure, and maintain a constant feature map resolution during network propagation. As the size of the network receptive field is limited, the pixel value inference of each spatial location only relies on small local surrounding regions, it is usually arduous to remove longer rain streaks~(e.g. third row of Fig.~\ref{fig:teaser}). Moreover, due to the ignorance of long-distance spatial context modeling, these models often have difficulty in accurately filling raindrop-removed image content while detecting heavy rain streaks, resulting in an often overly blurred result, especially on texture-rich edges~(e.g. first row in Fig.~\ref{fig:teaser}). Although various deep CNN based solutions have been proposed, existing efforts either focus on the entrance of the networks by decomposing the input image into high and low frequency information~\cite{ddn} or design cascaded learning schemes to decompose the task of rain removal into multi-stages~\cite{JRDR,DID-MDN}. A contextualized dilated network is proposed in~\cite{JRDR} to aggregate context information from three scales of receptive filed for more effective rain streak feature learning. All of these methods use  convolutional neural network as an encapsulated end-to-end mapping module without deepening into the rationality and superiority of neural network design towards more effective rain streaks removal. 

Inspired by the adaptive nonlocal means filter~\cite{de-rain_nonlocal_mean} for efficient single-image rain streaks removal, we proposed to incorporate non-local operation~\cite{NonLocal} to the design of our end-to-end de-raining network framework. The non-local operation computes the feature response at a spatial position as a weighted sum of the features at a specific range of positions in the considered feature maps~\cite{NonLocal}. Specifically, we propose a non-locally enhanced encoder-decoder network framework for single-image de-raining. The core architecture of our trainable de-raining engine is a concatenation of an encoder network and a corresponding decoder network. It is designed to be a symmetrical structure and both the encoder and the decoder network are composed of three cascaded non-locally enhanced dense blocks (abbr.~NEDB). Each NEDB is designed as a residual learning module which contains a non-local feature map weighting followed by four densely connected convolution layers for hierarchical feature encoding and another convolution layer for residual inference. Moreover, we introduce the pooling striding mechanism in our encoder network to learn increasingly abstract feature representation, which results in a decrease in resolution with the enlargement of receptive filed. We further incorporate pooling indices computed in the max-pooling step of the corresponding encoder to perform non-linear upsampling in our decoder, which helps to preserve the structure and details in the resulted image. 

In summary, this paper has the following contributions:
\begin{itemize}
\item We propose a non-locally enhanced encoder-decoder network framework for single-image rain streaks removal. It is able to learn increasingly abstract feature representation for more accurate rain streaks modeling in the encoding stage, and can remove all levels of rain streaks while promisingly preserving the texture details during decoding. 
\item We propose to incorporate a non-locally enhanced dense block~(NEDB) in our encoder-decoder framework, which can not only fully exploit hierarchical features from all embedded convolutional layers but also well capture the long-distance dependencies for spatial context modeling. 
\item Experimental results on both synthetic and real datasets have demonstrated the superiority of our proposed method, which achieves significant improvements over the state-of-the-art methods.
\end{itemize}

\begin{figure*}[t]
    \centering
    \begin{center}
        \includegraphics[width=1\linewidth]{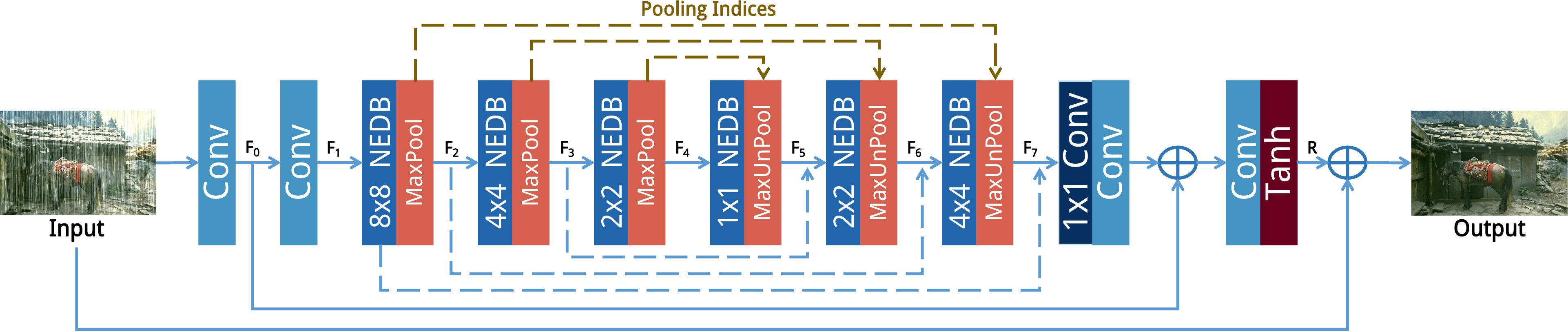} 
    \end{center}
    \caption{The overall architecture of our proposed non-locally enhanced encoder-decoder network (NLEDN). As can be observed, the input image and low-level feature activation are linked to the very end of the whole architecture via long-range skip-connections. The core of the whole architecture is a non-locally enhanced encoder-decoder, in which novel non-locally enhanced dense blocks (NEDBs) and pooling indices guided scheme are adopted.}
    \label{fig:architecture}
\end{figure*}

\begin{figure*}[t]
    \centering
    \begin{center}
        \includegraphics[width=0.8\linewidth]{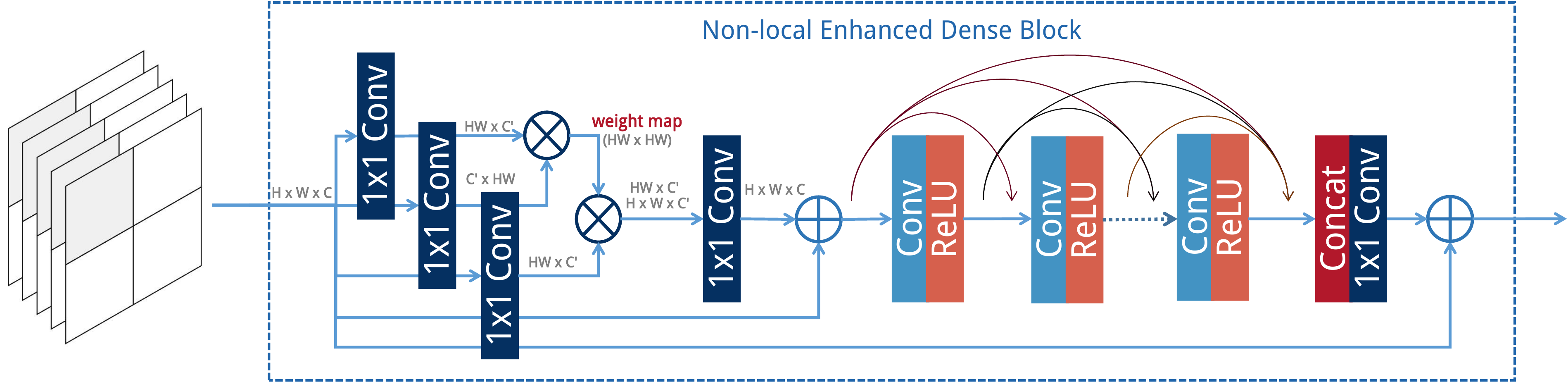} 
    \end{center}
    \caption{The architecture of our proposed non-locally enhanced dense block (NEDB). Left part shows the multi-scale input via either adopting global-level non-local enhancement which feeds the entire feature map to NEDB or dividing the feature map into a grid of regions to realize region-level non-local enhancement. Here we show by a $2\times2$ grid for convenience.}
    \label{fig:NEB}
\end{figure*}

\section{Related Works}
\subsection{Single Image De-raining}
Single image de-raining is a challenging and highly ill-posed task. Traditional approaches treat single image de-raining as an image decomposition problem, in which they model the rain streaks and rain-free scene lie in two separate sub-spaces. For example, Kang et al. \cite{sparse_coding_1} and Li et al. \cite{GMM} rely on morphological component analysis and Gaussian mixture models (GMMs) based dictionary learning, respectively. Yu et al. \cite{sparse_coding_3} distinguishes the dictionaries of rain streak and rain-free scene via discriminative sparse coding. Zhu et al. \cite{sparse_coding_4} further considers rain direction in a joint optimization process. In these methods, although varies prior information of both rain streaks and rain-free images have been extensively exploited, due to the handcraft low-level feature representation and strong prior assumptions, they usually tend to overly smooth the details in rain free scenes.

Recently, deep learning based approaches dominate the research of image-to-image mapping for various of computer vision tasks, such as image inpainting \cite{li2017context}, saliency detection \cite{li2018contrast,li2016visual,li2017instance,li2018flow}, automatic image colorization \cite{zhang2017automatic} and image super-resolution\cite{cao2017attention,SRCNN,RDN,zssr,SR_1,SR_2}. With the integration of several commonly used advances in network architectures such as residual connections \cite{Resnet} and dilated convolutions \cite{dilated_conv}, recent de-raining methods \cite{ddn,DID-MDN,JRDR} obtain impressive results by building fully convolutional architectures that learns pixel-wise mapping from rainy image to the rain-free version. Although additional considerations have been taken, such as rain density \cite{DID-MDN}, frequency domain knowledge \cite{ddn}, etc., they are still weak in removing long rain streaks in complex background scenes as well as distinguishing dense rain streaks with similar image patterns in rain-free ones. This is due to the fact that convolutions in existing CNN-based de-raining neural networks are inherently local operations with small range of spatial receptive field in each computation.

\subsection{Non-local Networks}
Very recently, non-local neural network is proposed to realize the computation of long-range dependencies \cite{NonLocal} for video classification task. In each 2D non-local operation, the response at a position is computed as a weighted sum of the features at all spatial positions. This is the first component of neural network that is able to enlarge the receptive field from neighbor positions to the entire image. Interestingly, such non-local operation is majorly inspired by traditional non-local mean filtering which is also exploited in early single image de-raining method \cite{de-rain_nonlocal_mean}. This verifies importance of the non-local enhancement in our proposed de-raining CNN engine. As far as we know, our proposed non-locally enhanced CNN architecture is the first piece of work that attempts to incorporate non-local mean calculation into a fully convolutional neural network architecture with correlation propagation for the task of pixel-wise image restoration.

\section{Method}\label{sec:method}
We introduce an end-to-end convolutional neural network for single image de-raining, called non-locally enhanced encoder-decoder network (NLEDN). Our framework contains a fully convolutional encoder-decoder network which has been proven able to learn complex pixel-wise mappings from large amount of input-output image pairs. The overall architecture of the proposed network is illustrated in Figure \ref{fig:architecture}. Particularly, in order to exploit the abundant structure cues in rain streak maps and the self-similarities in rain-free nature scenes, we propose the non-locally enhanced dense block (NEDB) as the basic component in our network architecture. We carefully integrate NEDBs with both encoding and decoding layers to enable the computation of long-range spatial dependencies as well as efficient usage of the feature activation of proceeding layers. In the following sections, we introduce each component of the proposed architecture with more detail.

\subsection{Entrance and Exit Layers}
The proposed de-raining NLEDN takes one image with rain-streaks as input in the entrance and outputs its rain-free version in the exit. In this section, we focus on the network structure of the entrance and exit of our entire framework. 

\noindent\textbf{Shallow Feature Extraction}
In the very beginning of the whole architecture, we use two convolution layers to extract shallow features of the input rainy image, as shown in Figure \ref{fig:architecture}. Formally, we have
\begin{equation}
\begin{aligned}
F_{0} = H_{0}(I_{0}),
\end{aligned}
\end{equation}
where $I_{0}$ and $H_{0}(\cdot)$ denote the input rainy image and the first shallow feature extraction convolution layer, respectively. As shown in Figure \ref{fig:architecture}, using the long-range skip-connections which bypass intermediate layers, we link both the input image $I_{0}$ and the shallow features $F_{0}$ with layers that are close to the exit of the whole network. The benefits of such skip-connections here are two folds: first, they provide long-range information compensation such that raw pixel values and low level feature activation are still available in the very end of the whole architecture; Second, they enable the residual learning that facilitates gradient back-propagation and the pixel-wise prediction.
Next, the shallow features $F_{0}$ is fed into the second convolution layer $H_{1}(\cdot)$ to obtain shallow features $F_{1}$:
\begin{equation}
\begin{aligned}
F_{1} = H_{1}(F_{0}),
\end{aligned}
\end{equation}
$F_{1}$ is used as the input to the subsequent encoding layers.

\noindent\textbf{Exit layers}
As shown in Figure \ref{fig:architecture}, the input image $I_{0}$ and the shallow features $F_{0}$ are gradually added to the feature activation of layers near the exit of the whole architecture. Particularly, one tanh layer is adopted as the nonlinear unit of the final convolution layer to obtain a rain map $R$ with pixel value within $(-1,1)$. The final rain-free image $\hat{Y}$ can be computed via
\begin{equation}
\begin{aligned}
\hat{Y} = I_{0} + R,
\end{aligned}
\end{equation}

Noted that the architecture of entrance and exit layers
here are not unique, we choose such architecture in order
to effectively incorporate our core modules, the non-locally
enhanced encoding and decoding layers, which will be elaborated in subsequent sections.

\subsection{Non-locally Enhanced Encoding and Decoding}
Conventional encoding-decoding networks are widely used in image-to-image translation or other pixel-wise prediction tasks. Here, the proposed architecture can be regarded as an enhanced version with non-local operations and dense connections. To achieve this, a novel non-locally enhanced dense block (NEDB) is plugged into each stage of both the encoder and the decoder.

\noindent\textbf{Non-locally Enhanced Dense Block (NEDB)}
The detailed architecture of the proposed NEDB is illustrated in Figure \ref{fig:NEB}. Specifically, we denote the input feature activation to the NEDB as $F_{n}$, which has the spatial dimension of $H_{n} \times W_{n} \times C_{n}$. 
The pair-wise function $f$ that calculates the pair-wise relationship is defined as

\begin{equation}
\begin{aligned}
f(F_{n,i},F_{n,j}) = \theta(F_{n,i})^{T}\phi(F_{n,j})
\end{aligned}
\end{equation}

where $F_{n,i},F_{n,j}$ denote the feature activation $F_{n}$ at position $i,j$ respectively. $\theta(\cdot)$ and $\phi(\cdot)$ are two feature embedding operations with different learned parameters $W_\theta$ and $W_\phi$, denoted as $\theta(F_{n,i})=W_\theta F_{n,i}$ and $\phi(F_{n,i})=W_\phi F_{n,i}$. Following 
\cite{NonLocal}, we define the non-local operation in NEDB as

\begin{equation}
\begin{aligned}
y_{n,i} = \frac{1}{C(F)} \sum_{\forall j} f(F_{n,i},F_{n,j})g(F_{n,j})
\end{aligned}
\end{equation}

where $g(\cdot)$ is the unary function that computes the representation of $F_{n}$ while $C(F)$ is the normalization factor, defined as $C(F)=\sum_{\forall j}f(F_{n,i}, F_{n,j})$. In this way, the feature representation is non-locally enhanced via considering all positions ($\forall j$) for each location $i$. 

Next, the non-locally enhanced feature activation is fed to five consecutive convolution layers which are densely connected. Specifically, following \cite{DenseNet}, we adopt direct connections from each layer to all subsequent layers. The architecture is shown in Figure \ref{fig:NEB}. Hence, the $l^{th}$ layer receives the feature activations from all preceding layers, $D_{0}\textrm{,...,}D_{l-1}$ as input:
\begin{equation}
\begin{aligned}
D_{l} = H_{l}([D_{0},...,D_{l-1}])
\end{aligned}
\end{equation}
where $[D_{0}\textrm{,...,}D_{l-1}]$ denotes the concatenation of the feature activations produced in layer $0 \textrm{,...,} l-1$.

Moreover, to avoid the notorious problem of gradients vanishing/exploding caused by an increase in the number of network layers and connections, we adopt local residual learning in the design of each NEDB. Formally, the final output of the $m$-th NEDB can be achieved by
\begin{equation}
\begin{aligned}
F_{m} = \mathcal{F}(F_{m-1},{W_{m}}) + F_{m-1},
\end{aligned}
\end{equation}
where $\mathcal{F}(F_{m-1},{W_{m}})$ represents the residual mapping to be learned in the considered block, which is actually inferred from a concatenation of feature activations from all preceding layers with a $1\times 1$ convolution layer, as illustrated in Fig.~\ref{fig:NEB} and formally written as 
\begin{equation}
\begin{aligned}
\mathcal{F}(F_{m-1},{W_{m}})=\conv_{1\times1}\left(\left[D_{0},...,D_{L}\right]\right).
\end{aligned}
\end{equation}

\noindent\textbf{Pooling Indices Guided Decoding}
As shown in Figure \ref{fig:architecture}, the encoding part consists of three consecutive NEDBs, each of which is followed by one max-pooling layer with striding that downsamples the feature activation. Symmetrically, another three NEDBs are stacked in the decoding part, each of which is followed by one max-unpooling layer that upsamples the feature activation. Moreover, skip-connections are utilized to link feature activations of encoding layers to their counterpart in the decoding layers. 

Particularly, we propose to record pooling indices~\cite{SegNet} during encoding for further upsample inferring in the decoding stage. The max-unpooling layer uses the pooling indices computed in the max-pooling step of the corresponding encoding layer to perform non-linear upsampling. Given the recorded pooling index matrix, the output feature map
of max-unpooling is calculated by first initializing to the size before the max pooling operation, and then assigning the feature column at each position of the input feature map to the corresponding position (given by the index matrix) and zeroing the remaining positions. The upsampled feature activations are sparse and convolved with subsequent trainable filters to produce dense feature maps. We will show by experiments that such pooling indices guided decoding scheme is more suitable here for rain streaks removal compared with conventional bilinear upsampling.

\noindent\textbf{Multi-scale Non-Local Enhancement}
With the above mentioned max-pooling and max-unpooling operations, the spatial size of intermediate feature activations gradually decreases in the encoding stage while gradually increases during decoding. Therefore, as the non-local operation in NEDB requires to compute pair-wise relations between every two spatial positions of the feature activation map, the computation burden increases dramatically when spatial dimension gets larger. To address this problem and construct a more flexible non-local enhancement across feature activations with different spatial resolution, we implement the non-local operation in a multi-scale manner when building the encoding and decoding layers.

Specifically, for the feature activation with lowest spatial resolution (e.g. F4 in Figure \ref{fig:architecture}), the subsequent NEDB directly works on the whole feature activation map to realize a global-level non-local enhancement. For the feature activation with higher spatial resolution, we first divide it into a grid of regions (As shown in Fig.~\ref{fig:architecture}, the $k\times k$ NEDB indicates how the input
feature map is divided before performing region-wise nonlocal
operation. e.g. $F_1$ is divided into a grid of $8\times8$) and then let the subsequent NEDB work on the feature activation inside each region. Accordingly, such region-level non-local enhancement is able to prevent unacceptable computational consumption caused by directly working on high resolution feature activation. One the other hand, comparing with conventional local convolution operation which operates inside $3\times3$ or $5\times5$ windows, region-level non-local enhancement is able to retrieve long-range structural cues to facilitate better rain-streaks removal. 

\subsection{Loss Function}
The loss function is defined as the mean absolute error~(MAE) between the resulted rain-free image and its corresponding groundtruth, which is formulated as follow:
\begin{equation}
\begin{aligned}
\mathcal{L} = \frac{1}{HWC} \sum_{i} \sum_{j} \sum_{k} \lVert \hat{Y}_{i, j, k} - Y_{i, j, k} \rVert_1,
\end{aligned}
\end{equation}
where $H$, $W$ and $C$ denotes the height, width and channel number of the rain-free image. $\hat{Y}$ and $Y$ denote the predicted rain-free image and ground-truth respectively.

\section{Experimental Results}
\subsection{Datasets and Evaluation Criteria}
We evaluate the performance of our proposed method (NLEDN) on both synthetic datasets and real data. For synthetic data, we use four benchmark datasets, including the dataset provided by Fu \textit{et al.}~\cite{ddn}, denoted as DDN-Data; the dataset synthesized by Zhang \textit{et al.}~\cite{DID-MDN}, denoted as DIDMDN-Data; the Rain100L and the Rain100H dataset provided in~\cite{JRDR}. Specifically, the DDN-Data contains 14,000 rainy/clean image pairs, which is synthesized from 1,000 clean images with 14 kinds of different rain-streak orientations and magnitudes. Following Fu \textit{et al.} \cite{ddn}, we select 9,100 pairs of them for training and the remaining 4,900 image pairs for evaluation. The DIDMDN-Data consists of 12,000 image pairs of three rain density levels~(i.e. light, medium and heavy). There are roughly 4,000 images per rain-density level in the dataset. The rain-density labels are also provided and are used as extra data for model training in~\cite{DID-MDN}. Noted that we did not use this extra information in our model. Rain100L is the synthesized dataset selected from BSD200~\cite{martin2001database} with only one type of rain streaks, which consists of 200 image pairs for training and the other 100 images for testing. Compared with Rain100L, Rain100H is more challenging. It is synthesized with five streaks directions and contains 1,800 images for training plus 100 images for testing. As pointed out in~\cite{JRDR}, although some of the synthesized examples in Rain100H are inconsistent with real images, adding these data as training can further enhance the robustness of the network. For real data, we collected some of the images from the Internet and some from the released images of~\cite{DID-MDN}. We have also taken some real cases using our own cameras for testing.

We evaluate the performance of the synthesized data using two metrics, including Peak Signal-to-Noise Ratio (PSNR)~\cite{huynh2008scope} and Structure Similarity Index (SSIM)~\cite{wang2004image}. As with existing works~\cite{JRDR,DID-MDN}, we evaluate the results in the luminance channel (i.e.~Y channel of YCbCr space), which has the most significant impact on the human visual system. As the rain-free groundtruth are not available on real-world image, we evaluate the performance on real data singly based on visual comparison.

\begin{table*}[ht]
    \centering
    \resizebox{0.95\textwidth}{!}{
\begin{tabular}{|c|c|c|c|c|c|c|c|}
            \hline
            Dataset & Metric & \multicolumn{1}{l|}{DSC \cite{sparse_coding_3} (ICCV'15)} & \multicolumn{1}{l|}{GMM \cite{GMM} (CVPR'16)} & DDN \cite{ddn} (CVPR'17) & JORDER \cite{JRDR} (CVPR'17) & DID-MDN \cite{DID-MDN} (CVPR'18) & Our NLEDN \\ \hline
            & PSNR & 22.03 & 25.64 & {\color[HTML]{32CB00} \textbf{28.24}} & {\color[HTML]{3531FF} \textbf{28.72}} & 26.17 & {\color[HTML]{FE0000} \textbf{29.79}} \\ \cline{2-8} 
            \multirow{-2}{*}{DDN-Data} & SSIM & 0.7985 & 0.8360 & {\color[HTML]{32CB00} \textbf{0.8654}} & {\color[HTML]{3531FF} \textbf{0.8740}} & 0.8409 & {\color[HTML]{FE0000} \textbf{0.8976}} \\ \hline
            & PSNR & 20.89 & 21.37 & 23.53 & {\color[HTML]{3531FF} \textbf{30.35}} & {\color[HTML]{32CB00} \textbf{*28.30}} & {\color[HTML]{FE0000} \textbf{33.16}} \\ \cline{2-8} 
            \multirow{-2}{*}{DIDMDN-Data} & SSIM & 0.7321 & 0.7923 & 0.7057 & {\color[HTML]{3531FF} \textbf{0.8763}} & {\color[HTML]{32CB00} \textbf{*0.8707}} & {\color[HTML]{FE0000} \textbf{0.9192}} \\ \hline
            & PSNR & 23.39 & 28.25 & 25.99 & {\color[HTML]{3531FF} \textbf{*35.23}} & {\color[HTML]{32CB00} \textbf{30.48}} & {\color[HTML]{FE0000} \textbf{36.57}} \\ \cline{2-8} 
            \multirow{-2}{*}{Rain100L} & SSIM & 0.8672 & 0.8763 & 0.8141 & {\color[HTML]{3531FF} \textbf{*0.9676}} & {\color[HTML]{32CB00} \textbf{0.9323}} & {\color[HTML]{FE0000} \textbf{0.9747}} \\ \hline
            & PSNR & 17.55 & 15.96 & 16.02 & {\color[HTML]{32CB00} \textbf{*25.21}} & {\color[HTML]{3531FF} \textbf{26.35}} & {\color[HTML]{FE0000} \textbf{30.38}} \\ \cline{2-8} 
            \multirow{-2}{*}{Rain100H} & SSIM & 0.5379 & 0.4180 & 0.3579 & {\color[HTML]{32CB00} \textbf{*0.8001}} & {\color[HTML]{3531FF} \textbf{0.8287}} & {\color[HTML]{FE0000} \textbf{0.8939}} \\ \hline
        \end{tabular}
    }
    \caption{Comparison of quantitative results in terms of PSNR and SSIM on four synthesized benchmark datasets. The three best performing algorithms are marked in \color[HTML]{FE0000}\textbf{red}\color{black}, \color[HTML]{3166FF}\textbf{blue}\color{black}, and \color[HTML]{32CB00}\textbf{green}\color{black}, respectively. Our proposed NLEDN consistently achieves the best performance. $`*'$ indicates that the method uses additional data~(e.g. rain density level, rain mask annotation) provided by the dataset.}
    \label{tab:comp-quantity}
\end{table*}

\subsection{Implementation}
Our proposed NLEDN has been implemented on the Pytorch framework, a flexible open source deep learning frame network. During training, we use horizontal flipping for data augmentation and resize the image to have long side smaller than 512. As the network is fully convolutional, and we set the mini-batch size to 1, the size of the input image does not have to be the same. We use adam optimizer to update the parameters of network during training. The learning rate is initially set to 0.0005, and we reduce it by 10\% when the training loss stops decreasing, until 0.0001. We use a weight decay of 0.0001 and a momentum of 0.9. Because of the large differences in the size of each datasets, the time spent on training each specific model is different. It takes around 3.5 days to train a model using the training set of DDN-Data or DIDMDN-Data, and it cost around 15 hours for training on Rain100H dataset. For Rain100L dataset, it is much faster and  only takes about four hours to complete a whole model training. Therefore, we conduct ablation study on Rain100L dataset in our experiment. However, as our entire model is fully convolutional, the testing process is very efficient, which only takes 1.44 seconds for the trained model to process a testing image with 512 $\times$ 512 pixels on a PC with an NVIDIA X GPU and a 3.4GHz Intel processor.

\begin{figure*}[ht]
\begin{center}
   \includegraphics[width=0.95\textwidth]{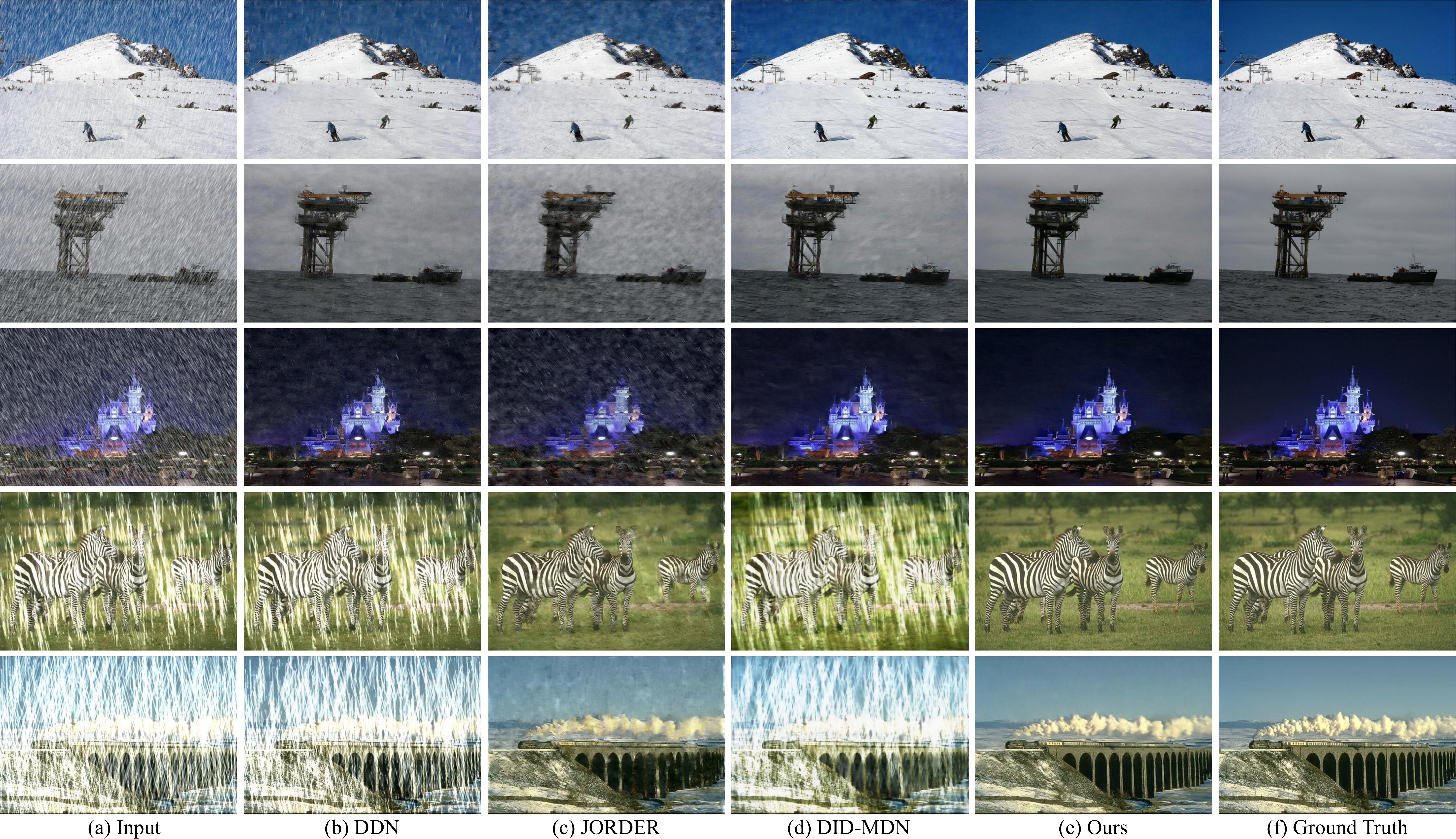}
\end{center}
   \caption{Visual comparison of rain-streaks removal results generated from state-of-the-art deep learning based methods (including our NLEDN) on synthesized rainy images. Our model consistently achieves the best visualization results in terms of effectively removing the rain streaks while preserving the image structure details. 
   }
\label{fig:visual_synthetic}
\end{figure*}

\begin{figure*}[ht]
\begin{center}
   \includegraphics[width=0.95\textwidth]{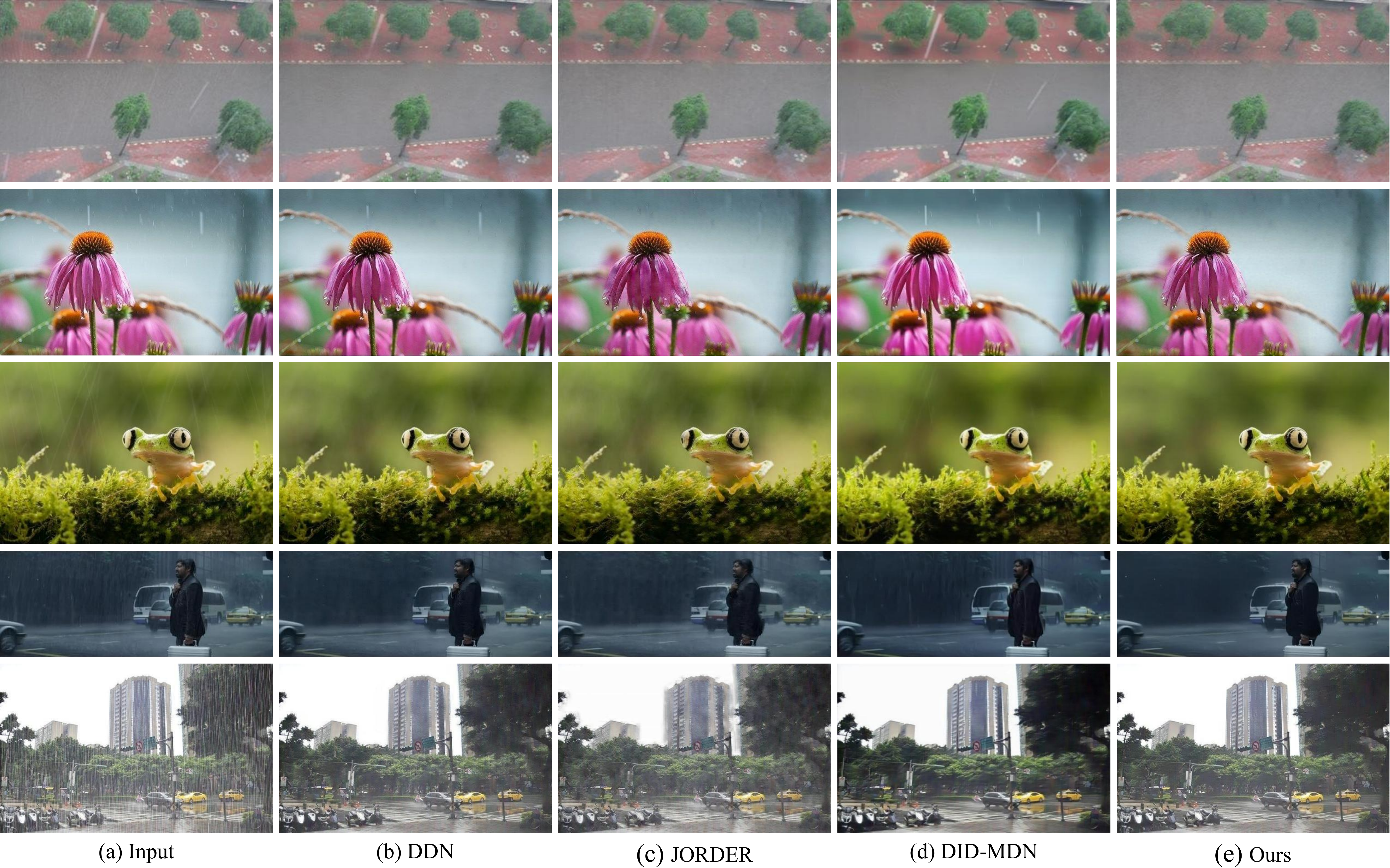}
\end{center}
   \caption{Visual comparison of rain-streaks removal results generated from state-of-the-art methods on real-world rainy images. Our model consistently achieves the best visualization results. 
   }
\label{fig:visual_real}
\end{figure*}

\begin{figure}[h]
\begin{center}
   \includegraphics[width=0.48\textwidth]{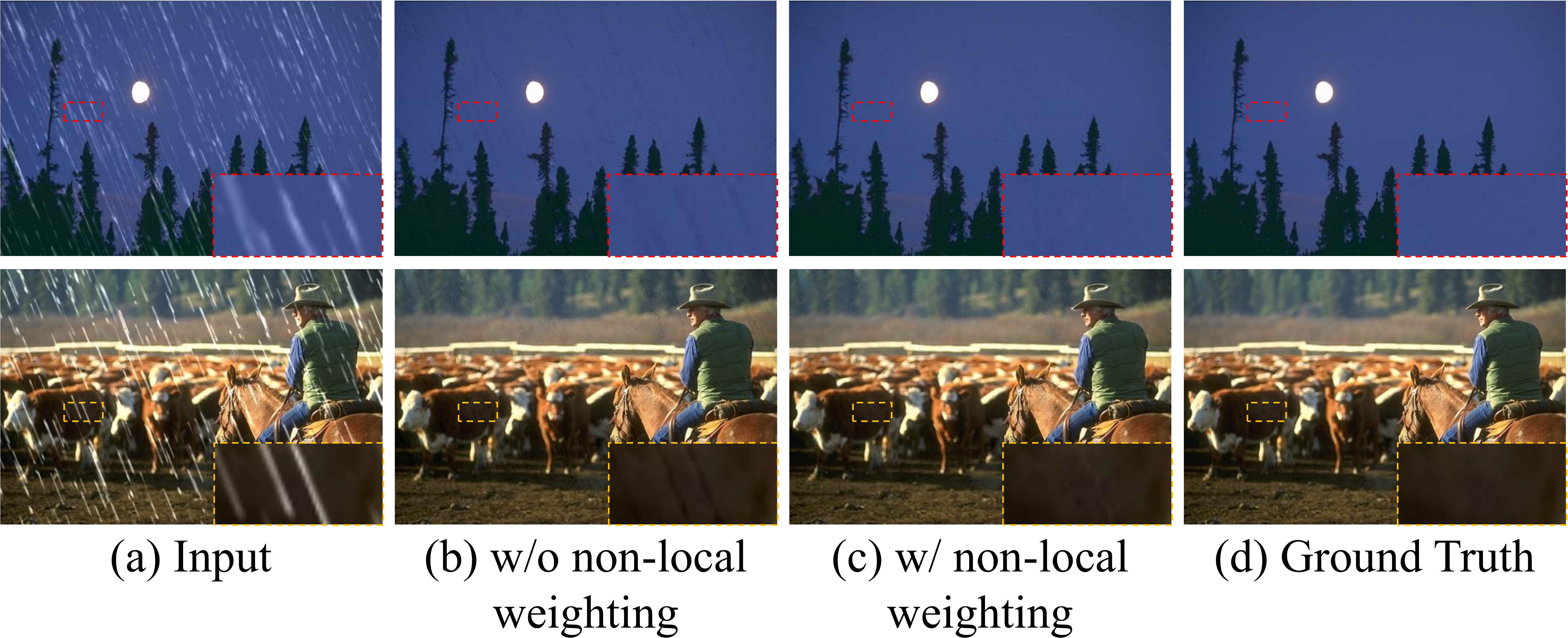}
\end{center}

   \caption{Visual comparison of rain-streaks removal results with and without non-local weighting enhancement in our proposed model.}
\label{fig:ablation_nonlocal}
\end{figure}

\subsection{Comparison with the state-of-the-art}
We compare our proposed NLEDN method against five state-of-the-art single-image de-raining methods, including discriminative sparse coding (DSC)~\cite{sparse_coding_3}, GMM-based layer prior (GMM)~\cite{GMM}, deep detail network (DDN)~\cite{ddn}, joint rain detection and removal (JORDER) \cite{JRDR} and density-aware single image de-raining using a multi-stream dense network (DID-MDN)~\cite{DID-MDN}. The last three are the latest deep learning based methods. As all of these methods use different data in training their models, we trained four versions of our model based on the training set of the four synthesized dataset respectively. Moreover, for fair comparison, we have also fine-tuned the released deep model of the comparison methods on each specific training set before evaluation. When evaluating on real-world data, we test four versions of the models for each deep learning based methods~(including our NLEDN) on each image and select the one with best visualization as the result for comparison. 

\textbf{Quantitative Evaluation.} We report a quantitative comparison w.r.t PSNR and SSIM in Table~\ref{tab:comp-quantity}. As can be observed, our proposed method~(NLEDN) increases the PSNR metric achieved by the existing best-performing algorithms by an average of 1.07db, 2.81db, 1.34db and 4.03db respectively on DDN-Data, DIDMDN-Data, Rain100L and Rain100H. And at the same time, our model improves the SSIM by 2.70\%, 4.90\%, 0.73\% and 7.87\% respectively on the above four datasets. We can find that, on average, the rain-free images restored by our NLEDN are consistently closer to the groundtruth than existing state-of-the-art on the synthesized datasets, and the higher SSIM value also indicates that our method can better restore the structural information of an image. Moreover, the more complex the data set, the more significant the performance of our algorithm compared to existing algorithms. Noted that JORDER \cite{JRDR} use additionally provided rain mask and rain-streak annotation while training their models on Rain100L and Rain100H datasets and DID-MDN \cite{DID-MDN} use extra rain density level information while training on their  DIDMDN-Data. Nonetheless, our proposed model can still greatly outperform their results without resorting to any additional data. 

\textbf{Qualitative Evaluation.} Fig.~\ref{fig:visual_synthetic} shows visual comparisons of rain-streaks removal results for five synthesized rainy images. The first three examples are synthetic heavy rain cases which are similar to real-world scenes, our models consistently achieves the best visualization results in terms of effectively removing the rain streaks while preserving the image structure details. The latter two are hard examples chosen from the Rain100H dataset, and may be rare in real world. As can be observed, both DDN~\cite{ddn} and DID-MDN~\cite{DID-MDN} are fail to remove rain streaks of these extreme cases though having been fine-tuned on this dataset. Although JORDER~\cite{JRDR} is designed to handle such hard cases, the restored images are over-smoothing and full of artifacts. Our proposed NLEDN generates much cleaner results with promisingly structure details preserving as it is able to fully exploit the long-range spatial context information based on gradually increased size of receptive field and the non-locally weighting scheme. Fig.~\ref{fig:visual_real} demonstrates the results of some real images with rain-streaks in various rain densities. As can be observed, our proposed model shows the best visual performance on rain-streaks removal. It is particularly effective in removing long rain-streaks while perfectly preserving the image structural details. Though the existing best-performing DID-MDN~\cite{DID-MDN} can remove long rain-streaks of high density to some extent, it suffers from artifacts and blurry if observed in a zoom-in view.

\begin{table*}[]
    \centering
    \begin{tabular}{c|c|c|c|c|c|c}
        \hline
        Methods & $R_a$ & $R_b$ & $R_c$ & $R_d$ & $R_e$ & $R_f$ \\ \hline
        without dense connection? & $\checkmark$ &  &  &  &  &  \\
        single block? & $\checkmark$ & $\checkmark$ &  &  &  &  \\
        multiple blocks? &  &  & $\checkmark$ & $\checkmark$ & $\checkmark$ & $\checkmark$ \\
        pooling striding? &  &  &  & $\checkmark$ &  & $\checkmark$ \\
        pooling indices? &  &  &  & $\checkmark$ &  & $\checkmark$ \\
        non-local operation? &  &  &  &  & $\checkmark$ & $\checkmark$ \\ \hline
        PSNR & 33.23 & 33.82 & 35.44 & 35.80 & 35.91 & 36.57 \\ \hline
        SSIM & 0.9533 & 0.9596 & 0.9691 & 0.9716 & 0.9720 & 0.9747 \\ \hline
    \end{tabular}
    \caption{Ablation study on different components of our proposed non-locally enhanced encoder-decoder network framework.}
    \label{tab:ablation}
\end{table*}

\subsection{Ablation Study}
As discussed in Section~\ref{sec:method}, our proposed NLEDN contains two core components towards more effective rain-streaks removal, including the encoder-decoder framework with pooling striding and max-unpooling operation, and the tailored non-locally enhanced dense block with densely connected convolution layers and a non-local feature weighting scheme. To validate the effectiveness and necessity of the internal network design of each of these two modules, we exhaustively compare NLEDN with its five variants trained and tested on the Rain100L dataset. The specific performance changes in terms of PSNR and SSIM are listed in Table~\ref{tab:ablation}. 

$R_a$ refers to the result of a very basic baseline which only includes the convolution layers of a single NEDB without dense connection or non-local weighting enhancement, as well as the same entrance and exit layer settings as NLEDN. Actually, it is a degenerate FCN based de-raining model with residual connection. It reaches the $PSNR = 33.23db$ and $SSIM = 0.9533$ which already outperforms the two recent deep learning based methods DDN~\cite{ddn} and DID-MDN~\cite{DID-MDN} but inferior to JORDER~\cite{JRDR}. $R_b$ adds dense connection between convolutional layers to $R_a$. As shown in the table, adding dense connection leads to an average increase of 0.59db in terms of PSNR and 0.7\% improvement on SSIM. This proves the effectiveness of dense connections. Due to space limitations,  in subsequent ablation studies, we add dense connection to each NEDB by default and discuss on the role of other network components.

$R_c$ is a simple concatenation of multiple dense blocks with neither non-local weighting nor receptive field controlling (pooling striding). For comparison, the number of blocks is set to the same as that of NLEDN. As shown in Table~\ref{tab:ablation}, simply concatenating multiple dense blocks can bring significant performance improvement, which increases the average PSNR by 1.62db while at the same time boosts the SSIM by 1\% when compared to its single block version $R_b$. This verifies the effectiveness of deeper feature representation in rain-streaks removal. In our experiments, we find that a cascade of 6 dense blocks leads to the best performance in our validation. Concatenating more than 6 dense blocks even leads to a performance deterioration. $R_d$ is directly modified on $R_c$ by adding pooling striding and corresponding skip connection between blocks guided by pooling indices. As illustrated in the table, adding a pooling indices based receptive filed controlling scheme further leads to an increase of 0.36db in terms of PSNR and a 0.3\% improvement on SSIM. 

$R_e$ and $R_f$ focus on the effectiveness of introducing non-local feature weighting to the network design. Firstly, we directly add multi-scale non-local enhancement to $R_c$ and observe a performance boost on PSNR from 35.44db to 35.91db and an increase of 0.3\% in terms of SSIM. The performance improvement is more significant when adding non-local operation to $R_d$, which forms our full model $R_f$ ($R_f$ = NLEDN). As shown in the table, the introduction of non-local operation contributes an average of 0.77db and 0.32\% improvement in terms of PSNR and SSIM respectively. This verifies the effectiveness and universality of non-local optimization for rain-streaks removal on single images. A visual comparison is provided in Fig.~\ref{fig:ablation_nonlocal}. As can be seen, the model without non-local enhancement suffers from some obvious artifacts and shows powerless for most of the long rain streaks. This further proves the effectiveness of long-distance dependencies modeling in rain-streaks removal.

\section{Conclusion}
In this paper, we have introduced a non-locally enhanced encoder-decoder network framework for rain streaks removal from single images. It is designed as a concatenation of an
encoder network followed by a corresponding decoder network, which are both composed of a series of tailored, non-locally enhanced dense blocks (NEDB). The NEDB is designed to not only fully exploit hierarchical features from densely connected convolutional layers but also well capture the long-distance dependencies and structural information by employing a non-locally weighting operation at a specific range of feature maps. Experimental results on both synthetic and real datasets have demonstrated that our proposed method can effectively remove rain-streaks on rainy image of various density while promisingly preserve the image texture similar to the rain streaks, which greatly outperforms the state-of-the-art. In our future research work, we plan to extend the proposed algorithm  to a wider range of image restoration tasks, including but not limited to image denoising, image dehazing and image super-resolution.

\bibliographystyle{ACM-Reference-Format}
\bibliography{derain}


\begin{thebibliography}{41}


\ifx \showCODEN    \undefined \def \showCODEN     #1{\unskip}     \fi
\ifx \showDOI      \undefined \def \showDOI       #1{#1}\fi
\ifx \showISBNx    \undefined \def \showISBNx     #1{\unskip}     \fi
\ifx \showISBNxiii \undefined \def \showISBNxiii  #1{\unskip}     \fi
\ifx \showISSN     \undefined \def \showISSN      #1{\unskip}     \fi
\ifx \showLCCN     \undefined \def \showLCCN      #1{\unskip}     \fi
\ifx \shownote     \undefined \def \shownote      #1{#1}          \fi
\ifx \showarticletitle \undefined \def \showarticletitle #1{#1}   \fi
\ifx \showURL      \undefined \def \showURL       {\relax}        \fi
\providecommand\bibfield[2]{#2}
\providecommand\bibinfo[2]{#2}
\providecommand\natexlab[1]{#1}
\providecommand\showeprint[2][]{arXiv:#2}

\bibitem[\protect\citeauthoryear{Badrinarayanan, Kendall, and
  Cipolla}{Badrinarayanan et~al\mbox{.}}{2017}]%
        {SegNet}
\bibfield{author}{\bibinfo{person}{V Badrinarayanan}, \bibinfo{person}{A
  Kendall}, {and} \bibinfo{person}{R Cipolla}.}
  \bibinfo{year}{2017}\natexlab{}.
\newblock \showarticletitle{SegNet: A Deep Convolutional Encoder-Decoder
  Architecture for Scene Segmentation.}
\newblock \bibinfo{journal}{\emph{IEEE Transactions on Pattern Analysis Machine
  Intelligence}} \bibinfo{volume}{PP}, \bibinfo{number}{99}
  (\bibinfo{year}{2017}), \bibinfo{pages}{2481--2495}.
\newblock


\bibitem[\protect\citeauthoryear{Bossu, Hautière, and Tarel}{Bossu
  et~al\mbox{.}}{2011}]%
        {video_derain_1}
\bibfield{author}{\bibinfo{person}{Jérémie Bossu}, \bibinfo{person}{Nicolas
  Hautière}, {and} \bibinfo{person}{Jean~Philippe Tarel}.}
  \bibinfo{year}{2011}\natexlab{}.
\newblock \showarticletitle{Rain or Snow Detection in Image Sequences Through
  Use of a Histogram of Orientation of Streaks}.
\newblock \bibinfo{journal}{\emph{International Journal of Computer Vision}}
  \bibinfo{volume}{93}, \bibinfo{number}{3} (\bibinfo{year}{2011}),
  \bibinfo{pages}{348--367}.
\newblock


\bibitem[\protect\citeauthoryear{Cao, Lin, Shi, Liang, and Li}{Cao
  et~al\mbox{.}}{2017}]%
        {cao2017attention}
\bibfield{author}{\bibinfo{person}{Qingxing Cao}, \bibinfo{person}{Liang Lin},
  \bibinfo{person}{Yukai Shi}, \bibinfo{person}{Xiaodan Liang}, {and}
  \bibinfo{person}{Guanbin Li}.} \bibinfo{year}{2017}\natexlab{}.
\newblock \showarticletitle{Attention-Aware Face Hallucination via Deep
  Reinforcement Learning}. In \bibinfo{booktitle}{\emph{Proceedings of the IEEE
  Conference on Computer Vision and Pattern Recognition}}.
  \bibinfo{pages}{1656--1664}.
\newblock


\bibitem[\protect\citeauthoryear{Chen and Hsu}{Chen and Hsu}{2014}]%
        {low_rank_1}
\bibfield{author}{\bibinfo{person}{Yi~Lei Chen} {and}
  \bibinfo{person}{Chiou~Ting Hsu}.} \bibinfo{year}{2014}\natexlab{}.
\newblock \showarticletitle{A Generalized Low-Rank Appearance Model for
  Spatio-temporally Correlated Rain Streaks}. In \bibinfo{booktitle}{\emph{IEEE
  International Conference on Computer Vision}}. \bibinfo{pages}{1968--1975}.
\newblock


\bibitem[\protect\citeauthoryear{Dong, Loy, He, and Tang}{Dong
  et~al\mbox{.}}{2016}]%
        {SRCNN}
\bibfield{author}{\bibinfo{person}{C. Dong}, \bibinfo{person}{C.~C. Loy},
  \bibinfo{person}{K. He}, {and} \bibinfo{person}{X. Tang}.}
  \bibinfo{year}{2016}\natexlab{}.
\newblock \showarticletitle{Image Super-Resolution Using Deep Convolutional
  Networks.}
\newblock \bibinfo{journal}{\emph{IEEE Transactions on Pattern Analysis Machine
  Intelligence}} \bibinfo{volume}{38}, \bibinfo{number}{2}
  (\bibinfo{year}{2016}), \bibinfo{pages}{295--307}.
\newblock


\bibitem[\protect\citeauthoryear{Fu, Huang, Zeng, Huang, Ding, and Paisley}{Fu
  et~al\mbox{.}}{2017}]%
        {ddn}
\bibfield{author}{\bibinfo{person}{Xueyang Fu}, \bibinfo{person}{Jiabin Huang},
  \bibinfo{person}{Delu Zeng}, \bibinfo{person}{Yue Huang},
  \bibinfo{person}{Xinghao Ding}, {and} \bibinfo{person}{John Paisley}.}
  \bibinfo{year}{2017}\natexlab{}.
\newblock \showarticletitle{Removing Rain from Single Images via a Deep Detail
  Network}. In \bibinfo{booktitle}{\emph{Proceedings of the IEEE Conference on
  Computer Vision and Pattern Recognition}}. \bibinfo{pages}{1715--1723}.
\newblock


\bibitem[\protect\citeauthoryear{Garg and Nayar}{Garg and Nayar}{2004}]%
        {video_derain_2}
\bibfield{author}{\bibinfo{person}{Kshitiz Garg} {and}
  \bibinfo{person}{Shree~K. Nayar}.} \bibinfo{year}{2004}\natexlab{}.
\newblock \showarticletitle{Detection and Removal of Rain from Videos}. In
  \bibinfo{booktitle}{\emph{Proceedings of the IEEE Conference on Computer
  Vision and Pattern Recognition}}.
\newblock


\bibitem[\protect\citeauthoryear{Garg and Nayar}{Garg and Nayar}{2005}]%
        {garg2005does}
\bibfield{author}{\bibinfo{person}{Kshitiz Garg} {and} \bibinfo{person}{Shree~K
  Nayar}.} \bibinfo{year}{2005}\natexlab{}.
\newblock \showarticletitle{When does a camera see rain?}. In
  \bibinfo{booktitle}{\emph{Proceedings of the IEEE International Conference on
  Computer Vision}}, Vol.~\bibinfo{volume}{2}. \bibinfo{pages}{1067--1074}.
\newblock


\bibitem[\protect\citeauthoryear{He, Zhang, Ren, and Sun}{He
  et~al\mbox{.}}{2016}]%
        {Resnet}
\bibfield{author}{\bibinfo{person}{Kaiming He}, \bibinfo{person}{Xiangyu
  Zhang}, \bibinfo{person}{Shaoqing Ren}, {and} \bibinfo{person}{Jian Sun}.}
  \bibinfo{year}{2016}\natexlab{}.
\newblock \showarticletitle{Deep Residual Learning for Image Recognition}. In
  \bibinfo{booktitle}{\emph{Proceedings of the IEEE Conference on Computer
  Vision and Pattern Recognition}}. \bibinfo{pages}{770--778}.
\newblock


\bibitem[\protect\citeauthoryear{Huang, Kang, Yang, Lin, and Wang}{Huang
  et~al\mbox{.}}{2012}]%
        {huang2012context}
\bibfield{author}{\bibinfo{person}{De-An Huang}, \bibinfo{person}{Li-Wei Kang},
  \bibinfo{person}{Min-Chun Yang}, \bibinfo{person}{Chia-Wen Lin}, {and}
  \bibinfo{person}{Yu-Chiang~Frank Wang}.} \bibinfo{year}{2012}\natexlab{}.
\newblock \showarticletitle{Context-aware single image rain removal}. In
  \bibinfo{booktitle}{\emph{Proceedings of the IEEE International Conference on
  Multimedia and Expo}}. \bibinfo{pages}{164--169}.
\newblock


\bibitem[\protect\citeauthoryear{Huang, Liu, and Weinberger}{Huang
  et~al\mbox{.}}{2016}]%
        {DenseNet}
\bibfield{author}{\bibinfo{person}{Gao Huang}, \bibinfo{person}{Zhuang Liu},
  {and} \bibinfo{person}{Kilian~Q. Weinberger}.}
  \bibinfo{year}{2016}\natexlab{}.
\newblock \showarticletitle{Densely Connected Convolutional Networks}. In
  \bibinfo{booktitle}{\emph{Proceedings of the IEEE Conference on Computer
  Vision and Pattern Recognition}}.
\newblock


\bibitem[\protect\citeauthoryear{Huynh-Thu and Ghanbari}{Huynh-Thu and
  Ghanbari}{2008}]%
        {huynh2008scope}
\bibfield{author}{\bibinfo{person}{Quan Huynh-Thu} {and}
  \bibinfo{person}{Mohammed Ghanbari}.} \bibinfo{year}{2008}\natexlab{}.
\newblock \showarticletitle{Scope of validity of PSNR in image/video quality
  assessment}.
\newblock \bibinfo{journal}{\emph{Electronics letters}} \bibinfo{volume}{44},
  \bibinfo{number}{13} (\bibinfo{year}{2008}), \bibinfo{pages}{800--801}.
\newblock


\bibitem[\protect\citeauthoryear{Kang, Lin, and Fu}{Kang et~al\mbox{.}}{2012}]%
        {sparse_coding_1}
\bibfield{author}{\bibinfo{person}{L.~W. Kang}, \bibinfo{person}{C.~W. Lin},
  {and} \bibinfo{person}{Y.~H. Fu}.} \bibinfo{year}{2012}\natexlab{}.
\newblock \showarticletitle{Automatic single-image-based rain streaks removal
  via image decomposition}.
\newblock \bibinfo{journal}{\emph{IEEE Transactions on Image Processing A
  Publication of the IEEE Signal Processing Society}} \bibinfo{volume}{21},
  \bibinfo{number}{4} (\bibinfo{year}{2012}), \bibinfo{pages}{1742--1755}.
\newblock


\bibitem[\protect\citeauthoryear{Kim, Lee, Sim, and Kim}{Kim
  et~al\mbox{.}}{2014}]%
        {de-rain_nonlocal_mean}
\bibfield{author}{\bibinfo{person}{Jin~Hwan Kim}, \bibinfo{person}{Chul Lee},
  \bibinfo{person}{Jae~Young Sim}, {and} \bibinfo{person}{Chang~Su Kim}.}
  \bibinfo{year}{2014}\natexlab{}.
\newblock \showarticletitle{Single-image deraining using an adaptive nonlocal
  means filter}. In \bibinfo{booktitle}{\emph{IEEE International Conference on
  Image Processing}}. \bibinfo{pages}{914--917}.
\newblock


\bibitem[\protect\citeauthoryear{Kim, Sim, and Kim}{Kim et~al\mbox{.}}{2015}]%
        {video_derain_3}
\bibfield{author}{\bibinfo{person}{Jin~Hwan Kim}, \bibinfo{person}{Jae~Young
  Sim}, {and} \bibinfo{person}{Chang~Su Kim}.} \bibinfo{year}{2015}\natexlab{}.
\newblock \showarticletitle{Video Deraining and Desnowing Using Temporal
  Correlation and Low-Rank Matrix Completion}.
\newblock \bibinfo{journal}{\emph{IEEE Transactions on Image Processing}}
  \bibinfo{volume}{24}, \bibinfo{number}{9} (\bibinfo{year}{2015}),
  \bibinfo{pages}{2658--70}.
\newblock


\bibitem[\protect\citeauthoryear{Kurihata, Takahashi, Ide, Mekada, Murase,
  Tamatsu, and Miyahara}{Kurihata et~al\mbox{.}}{2005}]%
        {kurihata2005rainy}
\bibfield{author}{\bibinfo{person}{Hiroyuki Kurihata},
  \bibinfo{person}{Tomokazu Takahashi}, \bibinfo{person}{Ichiro Ide},
  \bibinfo{person}{Yoshito Mekada}, \bibinfo{person}{Hiroshi Murase},
  \bibinfo{person}{Yukimasa Tamatsu}, {and} \bibinfo{person}{Takayuki
  Miyahara}.} \bibinfo{year}{2005}\natexlab{}.
\newblock \showarticletitle{Rainy weather recognition from in-vehicle camera
  images for driver assistance}. In \bibinfo{booktitle}{\emph{Proceedings of
  the IEEE Intelligent Vehicles Symposium}}. \bibinfo{pages}{205--210}.
\newblock


\bibitem[\protect\citeauthoryear{Lai, Huang, Ahuja, and Yang}{Lai
  et~al\mbox{.}}{2017}]%
        {SR_2}
\bibfield{author}{\bibinfo{person}{Wei-Sheng Lai}, \bibinfo{person}{Jia-Bin
  Huang}, \bibinfo{person}{Narendra Ahuja}, {and} \bibinfo{person}{Ming-Hsuan
  Yang}.} \bibinfo{year}{2017}\natexlab{}.
\newblock \showarticletitle{Deep laplacian pyramid networks for fast and
  accurate superresolution}. In \bibinfo{booktitle}{\emph{Proceedings of the
  IEEE Conference on Computer Vision and Pattern Recognition}}.
\newblock


\bibitem[\protect\citeauthoryear{Ledig, Wang, Shi, Theis, Huszar, Caballero,
  Cunningham, Acosta, Aitken, and Tejani}{Ledig et~al\mbox{.}}{2016}]%
        {SR_1}
\bibfield{author}{\bibinfo{person}{Christian Ledig}, \bibinfo{person}{Zehan
  Wang}, \bibinfo{person}{Wenzhe Shi}, \bibinfo{person}{Lucas Theis},
  \bibinfo{person}{Ferenc Huszar}, \bibinfo{person}{Jose Caballero},
  \bibinfo{person}{Andrew Cunningham}, \bibinfo{person}{Alejandro Acosta},
  \bibinfo{person}{Andrew Aitken}, {and} \bibinfo{person}{Alykhan Tejani}.}
  \bibinfo{year}{2016}\natexlab{}.
\newblock \showarticletitle{Photo-Realistic Single Image Super-Resolution Using
  a Generative Adversarial Network}.
\newblock  (\bibinfo{year}{2016}), \bibinfo{pages}{105--114}.
\newblock


\bibitem[\protect\citeauthoryear{Li, Xie, Lin, and Yu}{Li
  et~al\mbox{.}}{2017b}]%
        {li2017instance}
\bibfield{author}{\bibinfo{person}{Guanbin Li}, \bibinfo{person}{Yuan Xie},
  \bibinfo{person}{Liang Lin}, {and} \bibinfo{person}{Yizhou Yu}.}
  \bibinfo{year}{2017}\natexlab{b}.
\newblock \showarticletitle{Instance-level salient object segmentation}. In
  \bibinfo{booktitle}{\emph{Proceedings of the IEEE Conference on Computer
  Vision and Pattern Recognition}}. \bibinfo{pages}{247--256}.
\newblock


\bibitem[\protect\citeauthoryear{Li, Xie, Wei, Wang, and Lin}{Li
  et~al\mbox{.}}{2018}]%
        {li2018flow}
\bibfield{author}{\bibinfo{person}{Guanbin Li}, \bibinfo{person}{Yuan Xie},
  \bibinfo{person}{Tianhao Wei}, \bibinfo{person}{Keze Wang}, {and}
  \bibinfo{person}{Liang Lin}.} \bibinfo{year}{2018}\natexlab{}.
\newblock \showarticletitle{Flow Guided Recurrent Neural Encoder for Video
  Salient Object Detection}. In \bibinfo{booktitle}{\emph{Proceedings of the
  IEEE Conference on Computer Vision and Pattern Recognition}}.
  \bibinfo{pages}{3243--3252}.
\newblock


\bibitem[\protect\citeauthoryear{Li and Yu}{Li and Yu}{2016}]%
        {li2016visual}
\bibfield{author}{\bibinfo{person}{Guanbin Li} {and} \bibinfo{person}{Yizhou
  Yu}.} \bibinfo{year}{2016}\natexlab{}.
\newblock \showarticletitle{Visual saliency detection based on multiscale deep
  CNN features}.
\newblock \bibinfo{journal}{\emph{IEEE Transactions on Image Processing}}
  \bibinfo{volume}{25}, \bibinfo{number}{11} (\bibinfo{year}{2016}),
  \bibinfo{pages}{5012--5024}.
\newblock


\bibitem[\protect\citeauthoryear{Li and Yu}{Li and Yu}{2018}]%
        {li2018contrast}
\bibfield{author}{\bibinfo{person}{Guanbin Li} {and} \bibinfo{person}{Yizhou
  Yu}.} \bibinfo{year}{2018}\natexlab{}.
\newblock \showarticletitle{Contrast-Oriented Deep Neural Networks for Salient
  Object Detection}.
\newblock \bibinfo{journal}{\emph{IEEE Transactions on Neural Networks and
  Learning Systems}} (\bibinfo{year}{2018}).
\newblock


\bibitem[\protect\citeauthoryear{Li, Li, Lin, and Yu}{Li
  et~al\mbox{.}}{2017a}]%
        {li2017context}
\bibfield{author}{\bibinfo{person}{Haofeng Li}, \bibinfo{person}{Guanbin Li},
  \bibinfo{person}{Liang Lin}, {and} \bibinfo{person}{Yizhou Yu}.}
  \bibinfo{year}{2017}\natexlab{a}.
\newblock \showarticletitle{Context-Aware Semantic Inpainting}.
\newblock \bibinfo{journal}{\emph{arXiv preprint arXiv:1712.07778}}
  (\bibinfo{year}{2017}).
\newblock


\bibitem[\protect\citeauthoryear{Li, Tan, Guo, Lu, and Brown}{Li
  et~al\mbox{.}}{2016}]%
        {GMM}
\bibfield{author}{\bibinfo{person}{Yu Li}, \bibinfo{person}{Robby~T. Tan},
  \bibinfo{person}{Xiaojie Guo}, \bibinfo{person}{Jiangbo Lu}, {and}
  \bibinfo{person}{Michael~S. Brown}.} \bibinfo{year}{2016}\natexlab{}.
\newblock \showarticletitle{Rain Streak Removal Using Layer Priors}. In
  \bibinfo{booktitle}{\emph{Proceedings of the IEEE conference on computer
  vision and pattern recognition}}. \bibinfo{pages}{2736--2744}.
\newblock


\bibitem[\protect\citeauthoryear{Luo, Xu, and Ji}{Luo et~al\mbox{.}}{2015}]%
        {sparse_coding_3}
\bibfield{author}{\bibinfo{person}{Yu Luo}, \bibinfo{person}{Yong Xu}, {and}
  \bibinfo{person}{Hui Ji}.} \bibinfo{year}{2015}\natexlab{}.
\newblock \showarticletitle{Removing Rain from a Single Image via
  Discriminative Sparse Coding}. In \bibinfo{booktitle}{\emph{IEEE
  International Conference on Computer Vision}}. \bibinfo{pages}{3397--3405}.
\newblock


\bibitem[\protect\citeauthoryear{Martin, Fowlkes, Tal, and Malik}{Martin
  et~al\mbox{.}}{2001}]%
        {martin2001database}
\bibfield{author}{\bibinfo{person}{David Martin}, \bibinfo{person}{Charless
  Fowlkes}, \bibinfo{person}{Doron Tal}, {and} \bibinfo{person}{Jitendra
  Malik}.} \bibinfo{year}{2001}\natexlab{}.
\newblock \showarticletitle{A database of human segmented natural images and
  its application to evaluating segmentation algorithms and measuring
  ecological statistics}. In \bibinfo{booktitle}{\emph{Proceedings of the IEEE
  International Conference on Computer Vision}}, Vol.~\bibinfo{volume}{2}.
  \bibinfo{pages}{416--423}.
\newblock


\bibitem[\protect\citeauthoryear{Santhaseelan and Asari}{Santhaseelan and
  Asari}{2015}]%
        {santhaseelan2015utilizing}
\bibfield{author}{\bibinfo{person}{Varun Santhaseelan} {and}
  \bibinfo{person}{Vijayan~K Asari}.} \bibinfo{year}{2015}\natexlab{}.
\newblock \showarticletitle{Utilizing local phase information to remove rain
  from video}.
\newblock \bibinfo{journal}{\emph{International Journal of Computer Vision}}
  \bibinfo{volume}{112}, \bibinfo{number}{1} (\bibinfo{year}{2015}),
  \bibinfo{pages}{71--89}.
\newblock


\bibitem[\protect\citeauthoryear{Shen and Xue}{Shen and Xue}{2011}]%
        {shen2011fast}
\bibfield{author}{\bibinfo{person}{Minmin Shen} {and} \bibinfo{person}{Ping
  Xue}.} \bibinfo{year}{2011}\natexlab{}.
\newblock \showarticletitle{A fast algorithm for rain detection and removal
  from videos}. In \bibinfo{booktitle}{\emph{Proceedings of the IEEE
  International Conference on Multimedia and Expo}}. \bibinfo{pages}{1--6}.
\newblock


\bibitem[\protect\citeauthoryear{Shocher, Cohen, and Irani}{Shocher
  et~al\mbox{.}}{2018}]%
        {zssr}
\bibfield{author}{\bibinfo{person}{Assaf Shocher}, \bibinfo{person}{Nadav
  Cohen}, {and} \bibinfo{person}{Michal Irani}.}
  \bibinfo{year}{2018}\natexlab{}.
\newblock \showarticletitle{"Zero-Shot" Super-Resolution using Deep Internal
  Learning}. In \bibinfo{booktitle}{\emph{Proceedings of the IEEE Conference on
  Computer Vision and Pattern Recognition}}.
\newblock


\bibitem[\protect\citeauthoryear{Sun, Fan, and Wang}{Sun et~al\mbox{.}}{2014}]%
        {sun2014exploiting}
\bibfield{author}{\bibinfo{person}{Shao-Hua Sun}, \bibinfo{person}{Shang-Pu
  Fan}, {and} \bibinfo{person}{Yu-Chiang~Frank Wang}.}
  \bibinfo{year}{2014}\natexlab{}.
\newblock \showarticletitle{Exploiting image structural similarity for single
  image rain removal}. In \bibinfo{booktitle}{\emph{Proceedings of IEEE
  International Conference on Image Processing}}. \bibinfo{pages}{4482--4486}.
\newblock


\bibitem[\protect\citeauthoryear{Vasile, Vasile, Nistor, Vladareanu, Pantazica,
  Caldararu, Bonea, Drumea, and Plotog}{Vasile et~al\mbox{.}}{2010}]%
        {vasile2010rain}
\bibfield{author}{\bibinfo{person}{Alexandru Vasile}, \bibinfo{person}{Irina
  Vasile}, \bibinfo{person}{Adrian Nistor}, \bibinfo{person}{Luige Vladareanu},
  \bibinfo{person}{Mihaela Pantazica}, \bibinfo{person}{Florin Caldararu},
  \bibinfo{person}{Andreea Bonea}, \bibinfo{person}{Andrei Drumea}, {and}
  \bibinfo{person}{Ioan Plotog}.} \bibinfo{year}{2010}\natexlab{}.
\newblock \showarticletitle{Rain sensor for automatic systems on vehicles}. In
  \bibinfo{booktitle}{\emph{Advanced Topics in Optoelectronics,
  Microelectronics, and Nanotechnologies V}}, Vol.~\bibinfo{volume}{7821}.
  International Society for Optics and Photonics, \bibinfo{pages}{78211W}.
\newblock


\bibitem[\protect\citeauthoryear{Wang, Girshick, Gupta, and He}{Wang
  et~al\mbox{.}}{2018}]%
        {NonLocal}
\bibfield{author}{\bibinfo{person}{Xiaolong Wang}, \bibinfo{person}{Ross
  Girshick}, \bibinfo{person}{Abhinav Gupta}, {and} \bibinfo{person}{Kaiming
  He}.} \bibinfo{year}{2018}\natexlab{}.
\newblock \showarticletitle{Non-local Neural Networks}. In
  \bibinfo{booktitle}{\emph{Proceedings of the IEEE Conference on Computer
  Vision and Pattern Recognition}}.
\newblock


\bibitem[\protect\citeauthoryear{Wang, Bovik, Sheikh, and Simoncelli}{Wang
  et~al\mbox{.}}{2004}]%
        {wang2004image}
\bibfield{author}{\bibinfo{person}{Zhou Wang}, \bibinfo{person}{Alan~C Bovik},
  \bibinfo{person}{Hamid~R Sheikh}, {and} \bibinfo{person}{Eero~P Simoncelli}.}
  \bibinfo{year}{2004}\natexlab{}.
\newblock \showarticletitle{Image quality assessment: from error visibility to
  structural similarity}.
\newblock \bibinfo{journal}{\emph{IEEE transactions on image processing}}
  \bibinfo{volume}{13}, \bibinfo{number}{4} (\bibinfo{year}{2004}),
  \bibinfo{pages}{600--612}.
\newblock


\bibitem[\protect\citeauthoryear{Xue, Jin, Zhang, and Goto}{Xue
  et~al\mbox{.}}{2012}]%
        {xue2012motion}
\bibfield{author}{\bibinfo{person}{Xinwei Xue}, \bibinfo{person}{Xin Jin},
  \bibinfo{person}{Chenyuan Zhang}, {and} \bibinfo{person}{Satoshi Goto}.}
  \bibinfo{year}{2012}\natexlab{}.
\newblock \showarticletitle{Motion robust rain detection and removal from
  videos}. In \bibinfo{booktitle}{\emph{Proceedings of IEEE International
  Workshop on Multimedia Signal Processing}}. \bibinfo{pages}{170--174}.
\newblock


\bibitem[\protect\citeauthoryear{Yang, Tan, Feng, Liu, Guo, and Yan}{Yang
  et~al\mbox{.}}{2017}]%
        {JRDR}
\bibfield{author}{\bibinfo{person}{Wenhan Yang}, \bibinfo{person}{Robby~T.
  Tan}, \bibinfo{person}{Jiashi Feng}, \bibinfo{person}{Jiaying Liu},
  \bibinfo{person}{Zongming Guo}, {and} \bibinfo{person}{Shuicheng Yan}.}
  \bibinfo{year}{2017}\natexlab{}.
\newblock \showarticletitle{Deep Joint Rain Detection and Removal from a Single
  Image}. In \bibinfo{booktitle}{\emph{Proceedings of the IEEE Conference on
  Computer Vision and Pattern Recognition}}.
\newblock


\bibitem[\protect\citeauthoryear{Yu and Koltun}{Yu and Koltun}{2015}]%
        {dilated_conv}
\bibfield{author}{\bibinfo{person}{Fisher Yu} {and} \bibinfo{person}{Vladlen
  Koltun}.} \bibinfo{year}{2015}\natexlab{}.
\newblock \showarticletitle{Multi-Scale Context Aggregation by Dilated
  Convolutions}.
\newblock  (\bibinfo{year}{2015}).
\newblock


\bibitem[\protect\citeauthoryear{Zhang and Patel}{Zhang and Patel}{2018}]%
        {DID-MDN}
\bibfield{author}{\bibinfo{person}{He Zhang} {and} \bibinfo{person}{Vishal~M.
  Patel}.} \bibinfo{year}{2018}\natexlab{}.
\newblock \showarticletitle{Density-aware Single Image De-raining using a
  Multi-stream Dense Network}. In \bibinfo{booktitle}{\emph{Proceedings of the
  IEEE Conference on Computer Vision and Pattern Recognition}}.
\newblock


\bibitem[\protect\citeauthoryear{Zhang, Fang, and Li}{Zhang
  et~al\mbox{.}}{2017}]%
        {zhang2017automatic}
\bibfield{author}{\bibinfo{person}{Wei Zhang}, \bibinfo{person}{Chao-Wei Fang},
  {and} \bibinfo{person}{Guan-Bin Li}.} \bibinfo{year}{2017}\natexlab{}.
\newblock \showarticletitle{Automatic Colorization with Improved Spatial
  Coherence and Boundary Localization}.
\newblock \bibinfo{journal}{\emph{Journal of Computer Science and Technology}}
  \bibinfo{volume}{32}, \bibinfo{number}{3} (\bibinfo{year}{2017}),
  \bibinfo{pages}{494--506}.
\newblock


\bibitem[\protect\citeauthoryear{Zhang, Li, Qi, Leow, and Ng}{Zhang
  et~al\mbox{.}}{2006}]%
        {zhang2006rain}
\bibfield{author}{\bibinfo{person}{Xiaopeng Zhang}, \bibinfo{person}{Hao Li},
  \bibinfo{person}{Yingyi Qi}, \bibinfo{person}{Wee~Kheng Leow}, {and}
  \bibinfo{person}{Teck~Khim Ng}.} \bibinfo{year}{2006}\natexlab{}.
\newblock \showarticletitle{Rain removal in video by combining temporal and
  chromatic properties}. In \bibinfo{booktitle}{\emph{Proceedings of the IEEE
  International Conference on Multimedia and Expo}}. \bibinfo{pages}{461--464}.
\newblock


\bibitem[\protect\citeauthoryear{Zhang, Tian, Kong, Zhong, and Fu}{Zhang
  et~al\mbox{.}}{2018}]%
        {RDN}
\bibfield{author}{\bibinfo{person}{Yulun Zhang}, \bibinfo{person}{Yapeng Tian},
  \bibinfo{person}{Yu Kong}, \bibinfo{person}{Bineng Zhong}, {and}
  \bibinfo{person}{Yun Fu}.} \bibinfo{year}{2018}\natexlab{}.
\newblock \showarticletitle{Residual dense network for image super-resolution}.
  In \bibinfo{booktitle}{\emph{Proceedings of the IEEE Conference on Computer
  Vision and Pattern Recognition}}.
\newblock


\bibitem[\protect\citeauthoryear{Zhu, Fu, Lischinski, and Heng}{Zhu
  et~al\mbox{.}}{2017}]%
        {sparse_coding_4}
\bibfield{author}{\bibinfo{person}{Lei Zhu}, \bibinfo{person}{Chi~Wing Fu},
  \bibinfo{person}{Dani Lischinski}, {and} \bibinfo{person}{Pheng~Ann Heng}.}
  \bibinfo{year}{2017}\natexlab{}.
\newblock \showarticletitle{Joint Bi-layer Optimization for Single-Image Rain
  Streak Removal}. In \bibinfo{booktitle}{\emph{IEEE International Conference
  on Computer Vision}}. \bibinfo{pages}{2545--2553}.
\newblock


\end{thebibliography}

\end{document}